# Class Subset Selection for Transfer Learning using Submodularity

Varun Manjunatha
Dept. of Computer Science,
University of Maryland, College Park
varunm@cs.umd.edu

Srikumar Ramalingam
School of Computing,
University of Utah
ramalingam@cs.utah.edu

Tim Marks
Mitsubishi Electric Research Laboratories
tmarks@merl.com

Larry Davis
Dept. of Computer Science,
University of Maryland, College Park
lsd@umiacs.umd.edu

## Abstract

*In recent years, it is common practice to extract fully-connected-layer (fc) features that were learned while performing image classification on a source dataset, such as ImageNet, and apply them generally to a wide range of other tasks. The general usefulness of some large training datasets for transfer learning is not yet well understood, and raises a number of questions. For example, in the context of transfer learning, what is the role of a specific class in the source dataset, and how is the transferability of fc features affected when they are trained using various subsets of the set of all classes in the source dataset? In this paper, we address the question of how to select an optimal subset of the set of classes, subject to a budget constraint, that will more likely generate good features for other tasks. To accomplish this, we use a submodular set function to model the accuracy achievable on a new task when the features have been learned on a given subset of classes of the source dataset. An optimal subset is identified as the set that maximizes this submodular function. The maximization can be accomplished using an efficient greedy algorithm that comes with guarantees on the optimality of the solution. We empirically validate our submodular model by successfully identifying subsets of classes that produce good features for new tasks.*

## 1. Introduction

The following transfer learning scenario is now common in computer vision: Obtain a convolutional neural network that has been pretrained on a large data set for the task of classification. The last layer of this network is a softmax layer which corresponds to class probabilities. Upon removing this layer (and possibly one or more of the preceding fully connected layers), treat the remaining network as a generic feature extractor. For example, in the case of popular deep learning architectures such as AlexNet [20] or VGGNet [36], we can extract the activations of the hidden layer immediately before the classifier, or the previous hidden layer, to obtain a feature vector that describes an input image. These feature vectors are commonly referred to as *CNN features* or as *fc features* (fully connected features such as fc7 or fc6). For a new task, which may or may not be correlated to the original task, we use these fc features as the input to a new classifier (such as a linear SVM or a neural network) on a new dataset. Note that the classes in the new task may not match the original classes used in training the feature extractor.

This work addresses the following setting: A supervised image-classification task on one dataset (henceforth referred to as the *source task*) is used to learn general purpose features that can be used for classification tasks (referred to as the *target tasks*) on other datasets. In Figure 1 we consider the MNIST digit and Alphabet classification (or ALPHANIST)[13] problems as the source task and target task, respectively. Rather than training the classifier on the set of all 10 classes in the source task (the 10 digits $\{0, 1, \ldots, 9\}$), we consider training on a Subset Of Classes (SOC) from the source task (e.g., the set of 4 digits $\{0, 1, 4, 8\}$), then use the resulting fc feature extractors for the target task. As shown in the bar graph in Figure 1, each SOC leads to different a different generic feature extractor, which leads to a difference in classification accuracy on the target task. There seems to be a general assumption in the vision community that using all of the classes and training data from the source task leads to better fc features for the target task. (In our experiments, we define one set of fc features as *better* than a second set when the first set achieves higher classification accuracy on a target task.) We would

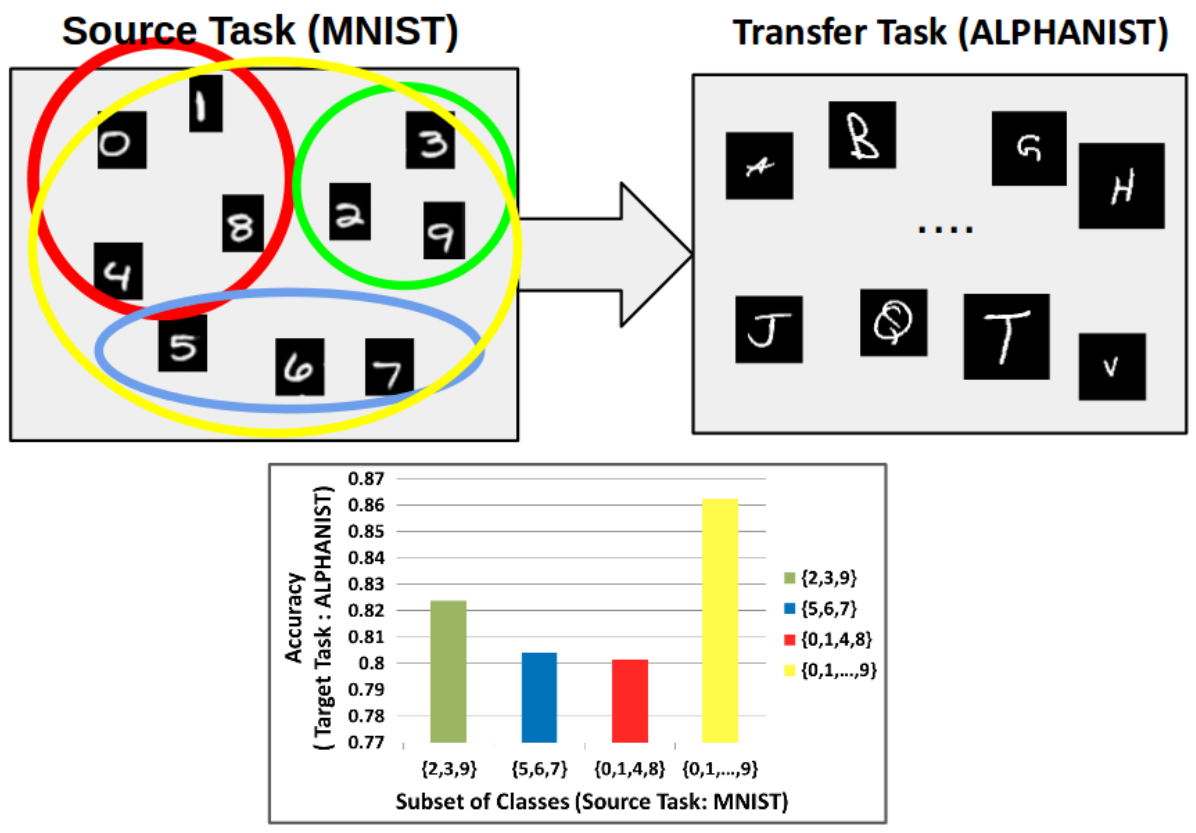

Figure 1. Illustration of the transfer learning scenario studied in this paper. In the upper left, we illustrate the source task (e.g., MNIST digit classification), in which each of several networks is trained using a different Subset of Classes (SOC) from the set of classes in the source task. In the upper right, we illustrate the target task (e.g., alphabet character classification), which is solved using the learned features from the source task. The bar graph shows the performance on the target task of networks that were trained on the source task using different SOCs. The goal of this work is a general method for identifying an Optimal SOC from the source task, under a budget constraint (a limited number of classes), to produce good generic features that obtain high accuracy on the target task. We refer to this problem as OPT-SOC.

like to investigate this assumption by asking the following questions:

1. **Monotonicity.** Does training on a larger number of classes from the source dataset always produce better features than training on a smaller number of classes? For example, in Figure 1, does a subset with 10 classes (e.g.,$\{0, \dots, 9\}$) always produce better features than a subset with 4 classes (e.g.,$\{0, 1, 4, 8\}$)?

2. **Optimal subset of classes.** Under a given budget (a fixed number of classes), what is the Optimal Subset Of Classes from a source dataset for generating good fc features? We refer to this problem as OPT-SOC. Without knowing much (if anything) about the target task, is it possible to know the optimal subset of a certain size that leads to the best fc features? In other words, can we determine which of two different subsets (e.g., $\{2, 3, 9\}$ versus $\{5, 6, 7\}$) will lead to better performance on the target task?

3. **Object class diversity:** Does it help or hurt the fc features when there are classes in the source dataset that are very similar (i.e., not diverse based on some similarity measure). For example, suppose the source dataset has classes that are visually similar, such as Leopard, Jaguar, and Cheetah. Would including more than one such class in a SOC (at the expense of a more diverse classes) lead to better-performing of generic fc features on a target task?

The problems addressed in this paper are challenging for two reasons. First, we do not assume any specific knowledge about the target task, so the selection of a SOC must be based entirely on the properties of the source dataset. Second, deep learning machinery for learning features from a specific data set is often seen as a "black box" that is still not completely understood, and thus identifying a pattern that can be generalized to many datasets is a significant challenge.

In this work, we model the function that maps a subset of classes in the source task to its end performance on a target task using submodular set functions (See Definition 2). Submodular functions are considered to be discrete analogues of convex functions, and they are generally used for modeling diminishing-return behavior in many learning problems. The diminishing return property states that the performance gain achieved by adding a class at an earlier stage is larger than that which would be obtained by adding it at a later stage. Using this modeling, we answer the three questions related to monotonicity, optimal subset selection, and object class diversity.

We summarize the contributions of this paper below:

- We propose a novel method that uses submodular set functions to model the performance of a set of generic features on unknown target tasks, as a function of the Subset Of Classes from a source task that was used to train the features.

- We propose two different ways to compute the parameters of the submodular set functions: (1) Linear programming for small-sized data sets; and (2) fitting a quadratic submodular function, based on a similarity score between pairs of classes, for large-scale data sets.

- We empirically show that our modeling allows us to find optimal subsets that performs significantly better on the target tasks than randomly selected subsets.

We envision three potential applications for our work:

- **Efficient training:** The proposed strategy for identifying the optimal Subset Of Classes in the source task leads to reduction in the size of the training data set, thereby enabling more efficient training of a generic fc feature extractor.

- **Data set generation:** If one wants to create a standard dataset for the sole purpose of generating generic fc feature extractors, this work can serve as a guideline for considering which combination of object classes would be most useful, thereby reducing the workload of manual annotation [30].

- **Better pre-trained models:** We observed that the full source dataset (the set consisting of all of the classes in the source dataset) does not always achieve the highest performance on the target task. OPT-SOC provides an efficient method to identify subsets that can potentially perform better than the set consisting of all the classes. This can be useful to compute better pre-trained models for transfer learning.

## 2. Related Work

**Subset selection in data sets:** Optimal subset selection is addressed by many researchers using submodular functions, which can be seen as a discrete analogue of convex functions. In particular, the formulation of the subset selection problem as the maximization of a submodular function has been used in many applications such as sensor placement [14], outbreak detection [23], word alignment [25], clustering [26], viral marketing [18], and finding diverse subsets in structured item sets [32].

Existing methods show that submodularity is generally well behaved in modeling many subset selection problems. However, OPT-SOC is different from existing problems. Existing subset selection problems typically perform dataset reduction by identifying a subset of training images and solve the same learning algorithm. In OPT-SOC, we don't use knowledge about the target task while finding the subset. The learning algorithm changes depending on the size of the chosen subsets. For example, for subsets of size two (2), our learning algorithm is a binary classification, whereas for subsets larger than 2, it is a multi-label classification. Furthermore, in OPT-SOC, the source and target tasks involve deep neural networks, which are highly non-convex and prone to local minima issues. Thus, it is not entirely obvious to see that OPT-SOC would benefit from submodular modeling. The main contribution in this paper it to model OPT-SOC using submodularity, and more importantly, to show that this can be beneficial in several data sets.

**Interpretations of CNN features:** In the last few years, there have been several papers in deep learning that achieve record-beating performance on challenging visual tasks such as image classification and object detection [20, 12, 36, 15]. From a scientific point of view, it would be useful to glean insight on these learned features, and a few recent papers address this. For example, by mapping the feature activations in intermediate layers to original input pixels using deconvolution, we can better understand the role of convnet features [43]. To interpret the features learned in a classification task, we can either generate an artificial image that is representative of a specific class of interest, or highlight the areas of an image that are discriminative of the object of interest [35]. It has been shown that while training convolutional neural networks for the task of scene recognition, the learned network also develops object detectors without using any explicit notion of objects [45]. Convolutional neural networks have been shown to have a few neurons that resemble the so-called "grandmother neuron" (a hypothetical neuron that is activated when presented with a specific object or concept), but most of neurons in convolutional neural networks form a distributed code [2].

A different line of research demonstrates that one can generate adversarial negatives, which introduce a small, hardly perceptible perturbation that leads to misclassification of an image. This raises an important concern regarding networks' ability to achieve high generalization performance [37]. Several other studies have looked at feature learning in the context of binarization [11], systematic variations in scene factors [4], variations in viewpoints [6], organization of class-specific information encoding [38, 39], egomotion [1, 17], temporal context [41], spatial context [10, 31, 29], and color [44]. Although there are many methods that show insights about the features learned by CNNs, we lack mathematical models to explain such phenomena [27].

**Understanding generic feature learning:** The problem of identifying the specific layers in a neural network that are suitable for transfer learning was studied in [42, 2]. In [33], the authors showed through extensive experiments that a linear SVM applied on generic 4096-dimensional fully connected features extracted from [20] could obtain or outperform state-of-the-art results on a wide variety of tasks. In their follow-up paper [5], the authors study a variety of factors that dictate the effectiveness of transfer learning. In a recent paper [16], the ImageNet dataset [34] is carefully studied to address several important questions: the relative importance of training samples, the relative importance of object classes, interaction between object classes, and comparison between limiting the number of classes versus limiting the number of training images per class. More interestingly, this work explicitly mentions the class subset selection as one of the interesting research questions to answer (the last sentence of Section 5.1 in [16]). The paper also reports observing the diminishing returns property on the target task with respect to the label set, without stating the connection to submodularity. In contrast to their work, we explicitly model the performance on the target task using a submodular set function. We also propose a method for identifying optimal subsets of classes top produce better features (a problem that is not addressed in [16]) and demonstrate the effectiveness of our method.

## 3. Notations and Preliminaries

In this paper we use a set function to denote the performance on a target task and we model this function to be submodular. Let $\mathbb{B}$ denote the Boolean set $\{0,1\}$ and $\mathbb{R}$ the set of reals. We use $\mathbf{x}$ to denote vectors.

**Definition 1.** *A set function $F : 2^E \rightarrow \mathbb{R}$, where $E$ is a finite set, maps a set to a real number. Set functions can also be seen as pseudo-Boolean functions [8] that take a Boolean vector as argument and return a real number.*

**Definition 2.** *A set function $F : 2^E \rightarrow \mathbb{R}$ is submodular if for all $A, B \subseteq E$ with $B \subseteq A$ and $e \in E \backslash A$, we have:*

$$F(A \cup \{e\}) - F(A) \leq F(B \cup \{e\}) - F(B). \quad (1)$$

This property is also referred to as diminishing return since the gain is less if the element $e$ is included at a later stage [28]. We analyse whether or not the performance of the target task improves by adding classes to a given subset. This behavior can be studied by analysing if the set function modeling the performance of the target task is monotonically increasing or not.

**Definition 3.** *A set function $F$ is monotonically increasing if for all $A, B \subseteq E$ and $B \subseteq A$, we have:*

$$F(B) \leq F(A) \quad (2)$$

## 4. Problem Statement

Let $\mathcal{C}$ denote the set of classes $\{c_1, c_2, \dots, c_n\}$ in the source task and let $\mathcal{T}$ denote a target task. We solve the source task by utilizing a subset of classes $A \subseteq \mathcal{C}$ and its associated training samples. The source task trains a generic feature extractor, and we use these features to solve the target task $\mathcal{T}$. Let $F_{\mathcal{T}} : 2^{\mathcal{C}} \rightarrow \mathbb{R}$ denote the accuracy or performance achieved on the target task $\mathcal{T}$ as shown below:

$$F_{\mathcal{T}}(A) = F(A) + C_{\mathcal{T}}, \forall A \subseteq \mathcal{C} \quad (3)$$

Here $F : 2^{\mathcal{C}} \rightarrow \mathbb{R}$ is independent of $\mathcal{T}$ and $C_{\mathcal{T}}$ is the task dependent constant. The basic assumption is that if a subset $A$ is better at solving a target task $\mathcal{T}_1$ in comparison to another subset $B$, then it is more likely that $A$ is also better at solving another transfer task $\mathcal{T}_2$ in comparison to the subset $B$. In other words, the source task produces a generic feature extractor that is independent of the target tasks. Our goal is to identify the optimal subset of classes, subject to a cardinality constraint, that produces good performance on the target task:

$$A^* = \arg\max_A F_{\mathcal{T}}(A) = \arg\max_A F(A), |A| \leq k, \quad (4)$$

where $k \leq |\mathcal{C}|$. It is important to note that the constant, which depends on the target task, does not affect the maximization, i.e., the selection of the optimal subset. We refer to $F : 2^{\mathcal{C}} \rightarrow \mathbb{R}$ as the transfer function and that is modelled as submodular set function and used to identify the optimal subset.

## 5. Algorithm

### 5.1. Submodular function modeling

We assume that the transfer function $F : 2^{\mathcal{C}} \rightarrow \mathbb{R}$ is a submodular set function. Let $x_i^A$ be a Boolean variable that indicates the presence of a class $c_i$ in a set $A \subseteq \mathcal{C}$, i.e., $x_i^A = 1$ if $c_i \in A$, and $x_i^A = 0$, if $c_i \notin A$. We denote the transfer function using the following quadratic Boolean function:

$$F(A) = \sum_{i=1}^{n} \alpha_i x_i^A + \sum_{i=1}^{n} \sum_{j=i+1}^{n} \beta_{ij} x_i^A x_j^A, \quad (5)$$

where $x_i^A$ and $x_j^A$ can be directly obtained from $A$ and $\beta_{ij} \leq 0$. The parameter $\alpha_i$ gives the role of the class $c_i$ in generating good features. The parameter $\beta_{ij}$ denotes the role of having two classes $c_i$ and $c_j$ jointly in the set $A$. Any quadratic pseudo-Boolean function with negative coefficients (i.e.,$\beta_{ij}$) for all bilinear terms is submodular [8]. This can be easily shown by checking the diminishing returns property for two sets $A, B \subseteq \mathcal{C}$ where $B \subseteq A$.

**Lemma 1.** *The function $F(A)$ is monotonically non-decreasing if $\alpha_i \geq -\sum_{c_j \in \mathcal{C} \backslash c_i} \beta_{ij}$*

See Section A for the proof.

We have shown a general form of submodular function for the transfer function in Equation 5 and the monotonicity conditions in Lemma 1. We will show two different ways to learn the parameters. In the first method, we use linear programming (LP) to compute solutions on the target task based on some SOC from the source dataset. Ideally, we would like to find the transfer function without using any information from the target task. In this paper, the LP is used to analyse the error in modeling the transfer function as a monotonically submodular function. The second method uses similarity matrix between pairs of classes in the source dataset, and this does not use any information from the target task.

**Parameter estimation using LP:** We would like to compute the parameters $\alpha$ and $\beta$ for a specific transfer learning setting. In order to do that, let us assume that we have some method to probe the value of the function $F_{\mathcal{T}}(A)$ on a target task $\mathcal{T}$ for different subsets $A \subseteq \mathcal{C}$. Note that the probed values would include an unknown task-specific constant term $C_{\mathcal{T}}$ as shown in Equation 3. Based on the probed values for different subsets, we fit a monotonically non-decreasing submodular function $F(A)$ for the transfer function by minimizing the sum of the $L_1$ norm distances between $F_{\mathcal{T}}(A) - C_{\mathcal{T}}$ and the fitted function for different probed values of $A \subseteq \mathcal{C}$. We propose an LP to compute the parameters of the transfer function as shown below:

$$\{\alpha, \beta\} = \arg\min_{\alpha,\beta} \sum_{A \subseteq \mathcal{C}} |s_A| \tag{6}$$

$$s.t$$
$$F_{\mathcal{T}}(A) = C_{\mathcal{T}} + F(A),$$
$$F(A) + s_A = \sum_{i=1}^{n} \alpha_i x_i^A + \sum_{i=1}^{n} \sum_{j=i+1}^{n} \beta_{ij} x_i^A x_j^A,$$
$$\alpha_i \geq -\sum_{c_j \in \mathcal{C} \setminus c_i} \beta_{ij}, \forall A \subseteq \mathcal{C}, \beta_{ij} \leq 0$$

Once we solve the $\alpha_i$ and $\beta_{ij}$ parameters using LP, we have the solution for $F$ using Equation 5. The cost function $\sum_{A \subseteq \mathcal{C}} |s_A|$ in the LP, which is the sum of the absolute values of the slack variables, gives some measure of how close the transfer function is to a monotonically non-decreasing submodular function. Note that the LP uses the probed values for the target task to fit the transfer function.

**Parameter estimation using similarity matrix:** The method to compute parameters using LP is computationally infeasible for data sets with large number of classes. For such scenarios, we propose an alternative method to compute the parameter $\beta_{ij}$, which denotes the interaction between two classes, using some measure of class similarity [40, 21, 7, 3]. In this paper, we use Wordnet tree to compute the similarity between pairs of classes. The critical assumption we make here, which holds frequently in practice, is that classes similar in terms of Wordnet similarity are visually similar. The nodes in the Wordnet tree represent classes. The Lin similarity (one of the many similarity measures in computational linguistics community) between two classes $c_i$ and $c_j$ is given below:

$$S(c_i, c_j) = \frac{2logP(L(c_i, c_j))}{logP(c_i) + logP(c_j)} \tag{7}$$

where L is the lowest node in the tree which is a common ancestor to both $c_1$ and $c_2$. Here the classes can be seen as concepts and $P(c_i)$ denotes the probability of a random word consumed by the concept $c_i$. The Lin similarity $S(c_i, c_j)$ varies from 0 to 1. For more details, we refer the reader to [24]. We use $\beta_{ij} = -S(c_i, c_j)$ and $\alpha_i = |\mathcal{C}|$. Since we don't have any information about the relative importance of different classes, we set all of them to the same value $\alpha_i = |\mathcal{C}|$ that ensures monotonicity of the transfer function.

### 5.2. Optimal subset selection

The use of greedy algorithm for maximizing submodular function is motivated by the following theorem:

**Theorem 1.** *[28] For maximizing monotonically non-decreasing submodular functions under a cardinality constraint, the optimality of the greedy algorithm is given by the following equation:*

$$f(A_{greedy}) \geq (1 - \frac{1}{e}) f(A_{OPT}), \tag{8}$$

*where* $f(\emptyset) = 0$.

We can observe that our transfer function satisfies $F(\emptyset) = 0$. We briefly outline the greedy algorithm to select optimal subsets of size $k$ [28].

1. Initialize $S = \emptyset$.
2. Let $s = \arg\max_{s' \in \mathcal{C}} F(S \cup \{s'\}) - F(S)$ such that $|S \cup \{s'\}| \leq k$.
3. If $s \neq \emptyset$ then $S = S \cup \{s\}$ and go to step 2.
4. $S$ is the required subset.

## 6. Experiments

We conducted several experiments to address questions related to monotonicity, optimal SOC from the source dataset, and object class diversity. We briefly explain the datasets, computation of the transfer function, the network architectures for the source and target tasks, and results obtained using OPT-SOC.

### 6.1. Datasets

We use five datasets : MNIST[22], ALPHANIST[13], CIFAR-10[19], STL-10[9] and CIFAR-100 for our experiments, all of which are standard, except for ALPHANIST [1]. We summarize the datasets used in source and target tasks in Table 1. We use the ALPHANIST dataset for a target task in the first experiment. In this experiment, we randomly sample a subset of 16800 digits (12000 for train and 4800 for test). We resize the images to 28x28 to match the dimensions of MNIST, and remove the classes that are similar to the ones in the source task. For example, we remove the alphabets "O" and "I" due to their to similarity with "0" and "1" digits in MNIST. The class labels of STL-10 are same as CIFAR-10, but the images are obtained from ImageNet. We resize the STL-10 images to 32 x 32 for compatibility with CIFAR-10/100.

### 6.2. Transfer function estimation

In each of the four experiments, we computed a monotonically non-decreasing submodular transfer function. For the first two experiments, we were able to compute the function using both LP and similarity matrix. Note that the LP-based fitting is primarily done to analyse modeling error. In all the experiments, we probe the performance on the target task for different SOCs of the source task. We refer to them as probed values.

[1] https://www.nist.gov/srd/nist-special-database-19

| Expt | Source Dataset | No. source classes | No. models trained | Target Dataset | No. target classes |
|---|---|---|---|---|---|
| 1 | MNIST (60k) | 10 | 1013 | ALPHANIST | 24 |
| 2 | CIFAR10 (50k) | 10 | 1013 | CIFAR100 | 100 |
| 3 | CIFAR100 (50k) | 100 | 181 | CIFAR10 | 10 |
| 4 | CIFAR100 (50k) | 100 | 181 | STL10 | 10 |

Table 1. Source and target datasets along with the number of classes and trained models. Note that in the case of MNIST and CIFAR-10, we exhaustively train all possible subsets (1013 in number), while we sample a random selection of 181 subsets in the case of CIFAR-100.

**Parameter estimation using LP:** For a source dataset with $n$ classes, there are $2^n - n - 1$ subsets (after removing $n$ singleton sets and the null set). In the case of MNIST and CIFAR-10, which both have 10 classes, we thus have $1024 - 10 - 1 = 1013$ subsets, while for CIFAR-100, which has 100 classes, we have $2^{100} - 100 - 1 \approx 1.26 \times 10^{30}$ subsets. We test the LP-based parameter estimation for the MNIST and CIFAR-10 datasets. The mean modeling error from the slack variables in the LP for the first experiment is $0.01169$, and $0.00815$ for the second experiment. The transfer function $F$ can vary from 0 to 1. This implies that the error in submodular function approximation is around $1\%$.

**Parameter estimation using similarity matrix:** We use Wordnet to obtain similarity matrix and thus, the associated transfer functions for all experiments except the first one (MNIST-ALPHANIST). This is because there is no well established similarity measure for comparing digits.

### 6.3. Network Architectures

For each source task, we train a deep convolutional neural network with the configuration shown in Figure 2. The CNNs are trained with a categorical cross-entropy loss with SGD, and an initial learning rate = 0.01. We drop the learning rate to 1/10th whenever the validation accuracy goes through a plateau. When the validation accuracy does not improve for 7 consecutive epochs, we use an early stopping criterion to cull the training thus avoiding overfitting. We only consider the model that has performed best on the validation set randomly partitioned from the training set, for the rest of the protocol. For the target task, we use the target dataset and extract features in the penultimate (i.e., fully connected) layer and feed them into a shallow neural network with one hidden layer. This shallow neural network is trained thrice and results averaged, to account for variations. A brief summary of these shallow models is provided in Figure 3. In many target tasks, it is a standard practice to use a simple classification algorithm such as nearest neighbor, a simpler neural network or SVM for solving the target tasks.

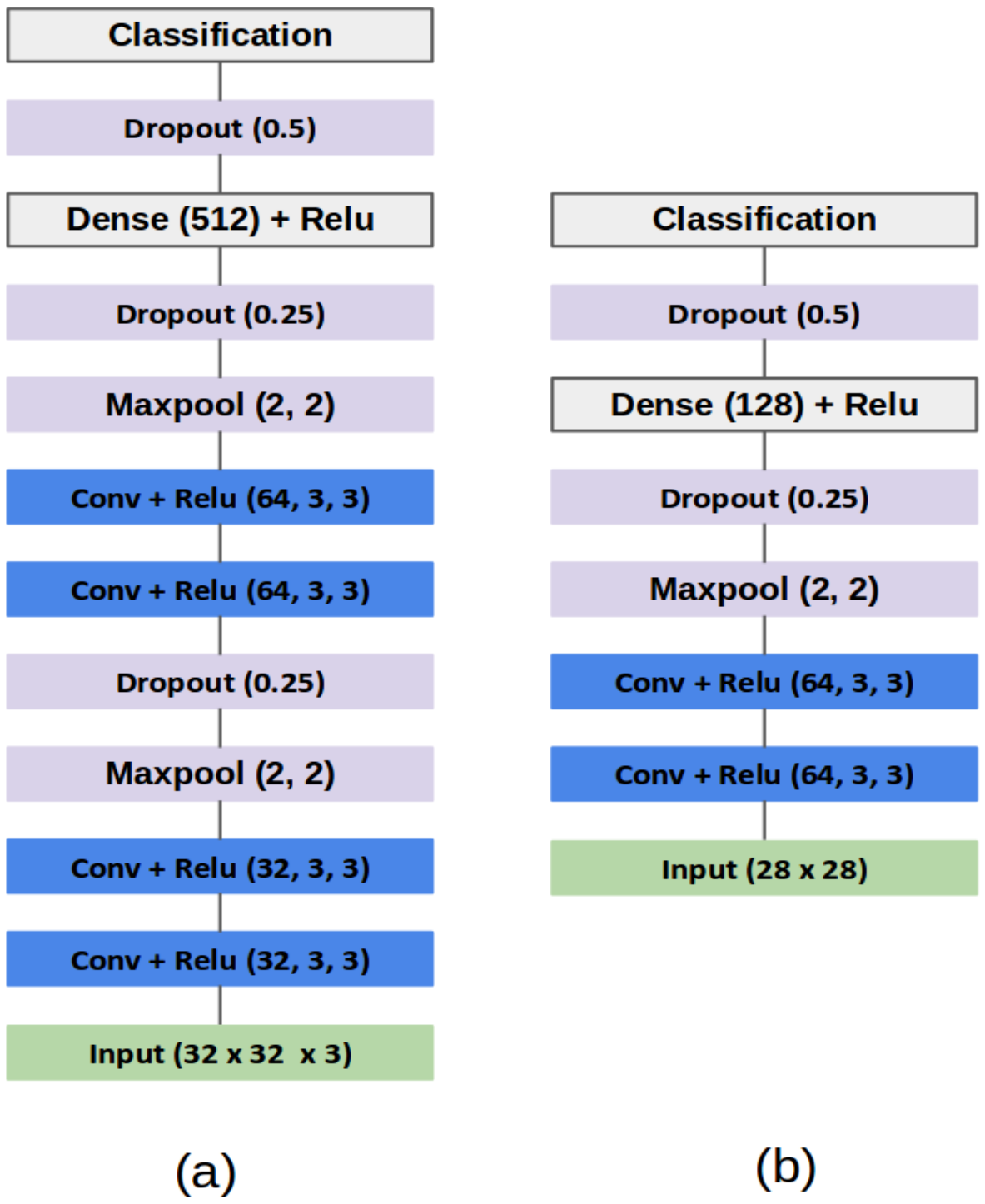

Figure 2. CNN architectures for source tasks. We use a standard CNN architecture with a cascade of convolution, max-pooling, fully connected, and dropout layers. The Dense(512/128) refers to the Fully Connected layer. The networks used for CIFAR and MNIST are shown in (a) and (b), respectively.

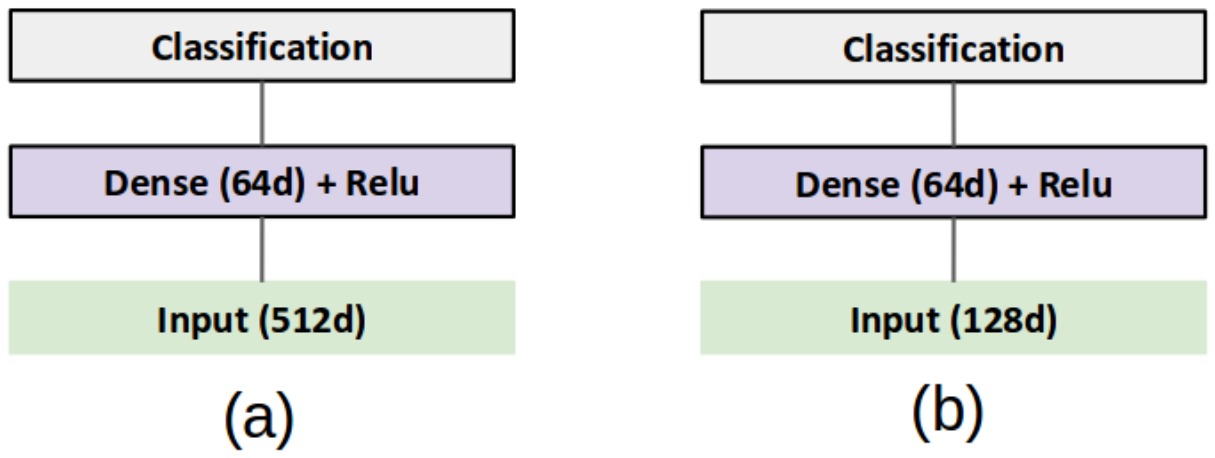

Figure 3. Neural network architectures for target tasks. We use a very simple neural network with one hidden layer of dimension 64 and a softmax classification. The networks used for CIFAR and MNIST are shown in (a) and (b), respectively.

### 6.4. OPT-SOC Computation

To evaluate the optimal SOC, we need the performance of the SOC on the target tasks. In the first two experiments, we use the 1013 probed values for all possible SOC from the source tasks. In the third and fourth experiments, we ran-domly sample 181 subsets (10 subsets each with cardinality $\{10, 15, 20, ...., 95\}$ and the complete dataset). We want to show that the optimal SOC, obtained using the greedy algorithm with the submodular transfer function $F$, performs better on the transfer tasks with respect to random subsets.

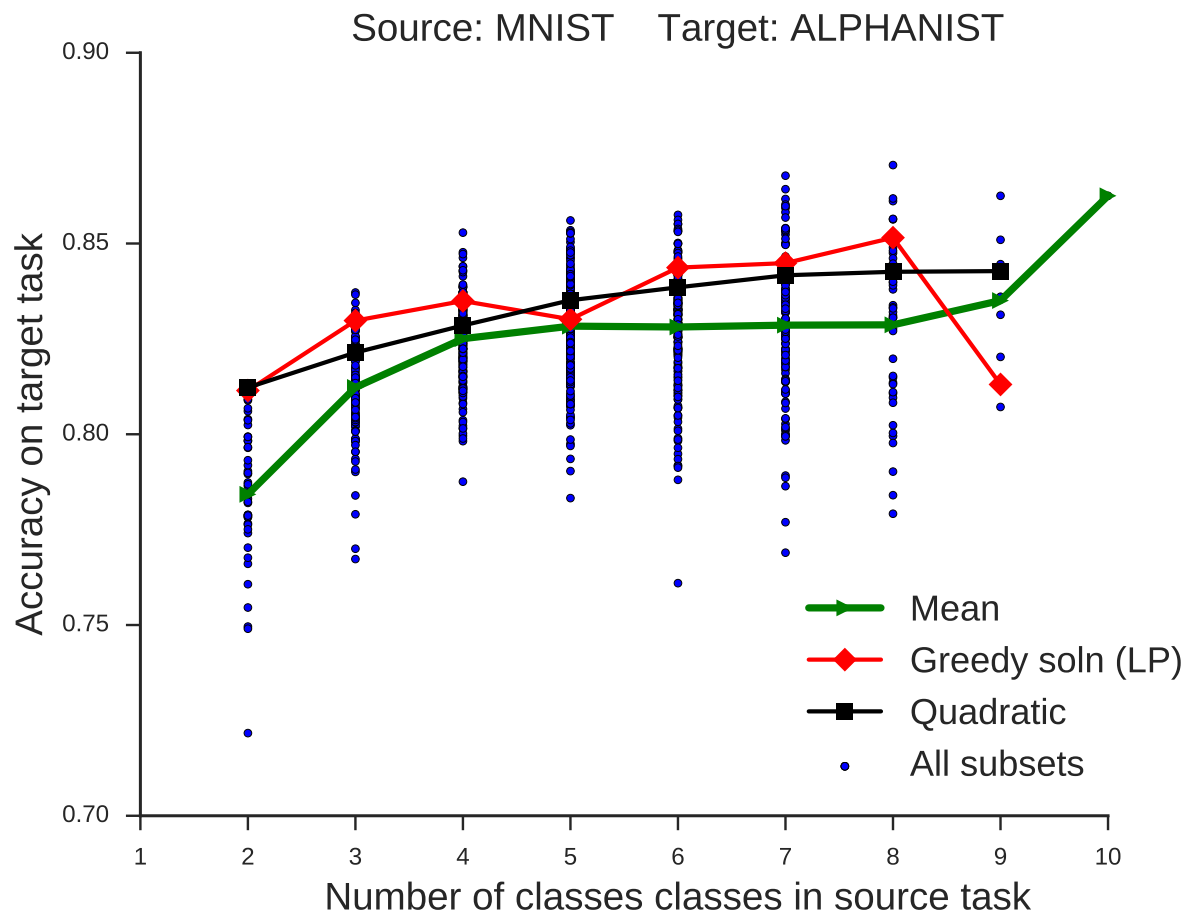

Figure 4. The fitted $F$ values (black) for the optimal SOC is a smooth monotonically non-decreasing curve. The Optimal SOC obtained using the greedy method (red) outperforms the baseline (green) for most of the cases.

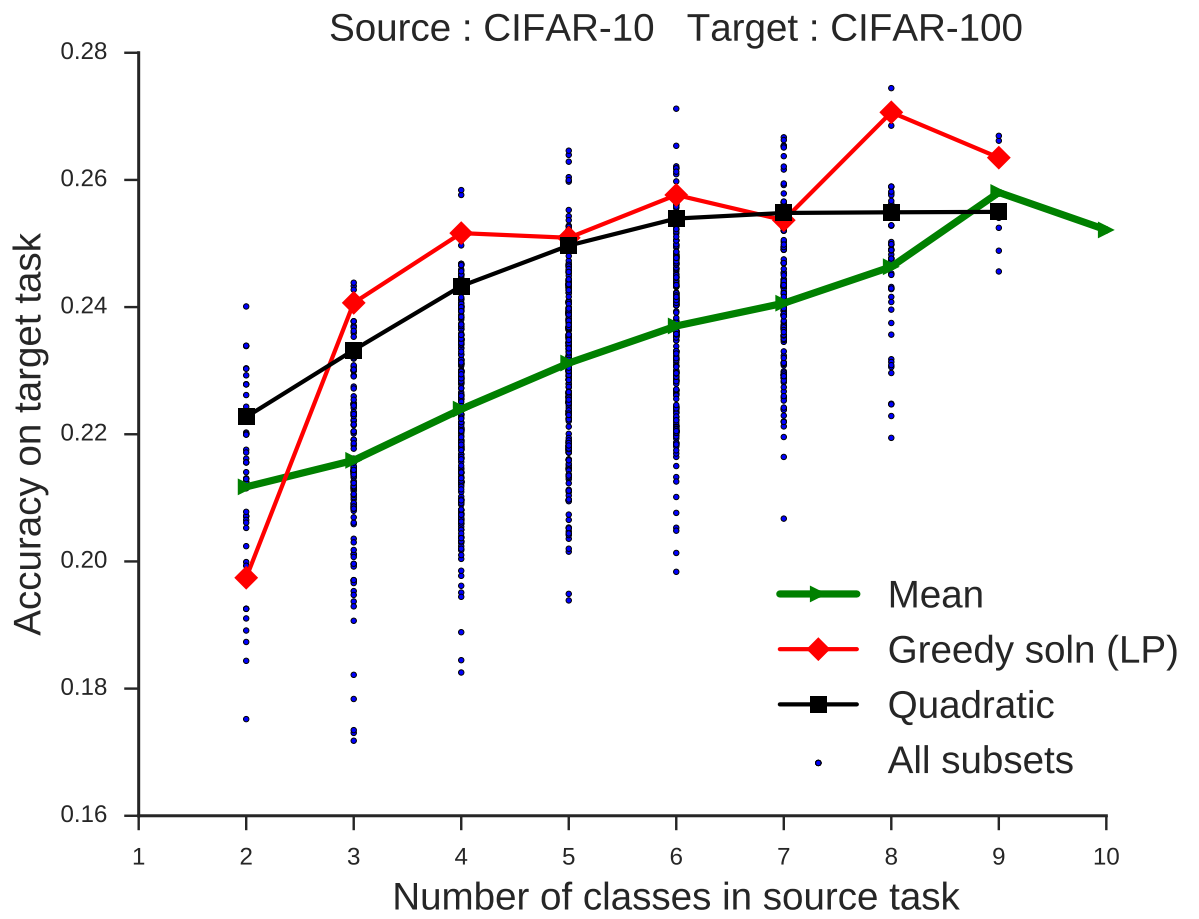

Figure 5. The fitted $F$ values (black) for the optimal SOC is a smooth monotonically non-decreasing curve. The Optimal SOC obtained using the greedy method (red) outperforms the baseline (green) for most of the cases.

Given a submodular function $F$ obtained through LP or similarity matrix, we can find optimal SOC in the source dataset using the greedy algorithm. We show the performance of the optimal SOC in different experiments as shown in Figures 4 - 8. In all these graphs, the blue dots show the probed values on the target tasks. We find the mean score for different subsets with same cardinalities. The green curve that connects these mean values will be treated as the baseline. The red curve in all the graphs shows the probed values based on the optimal SOC chosen using the greedy algorithm. In Figures 4 and 5, the red curve shows the performance of the optimal SOC, obtained using submodular function computed using LP. In these figures ( 4 and 5), the black curve shows the actual $F$ values corresponding to the optimal subsets, after including the task dependent constant $C_{\mathcal{T}}$ that is computed in the LP. In Figures 6, 7, and 8, the red curve shows the performance of the optimal SOC, using similarity matrix.

## 7. Discussion

We observed that the use of submodularity for identifying optimal SOC can be beneficial in practice. We briefly address the questions related to monotonicity, optimal SOC, and the class diversity.

- Monotonicity: In most of our experiments (Figure 4 - 7) we observed that in general adding more classes improves the performance on the target task. However, in

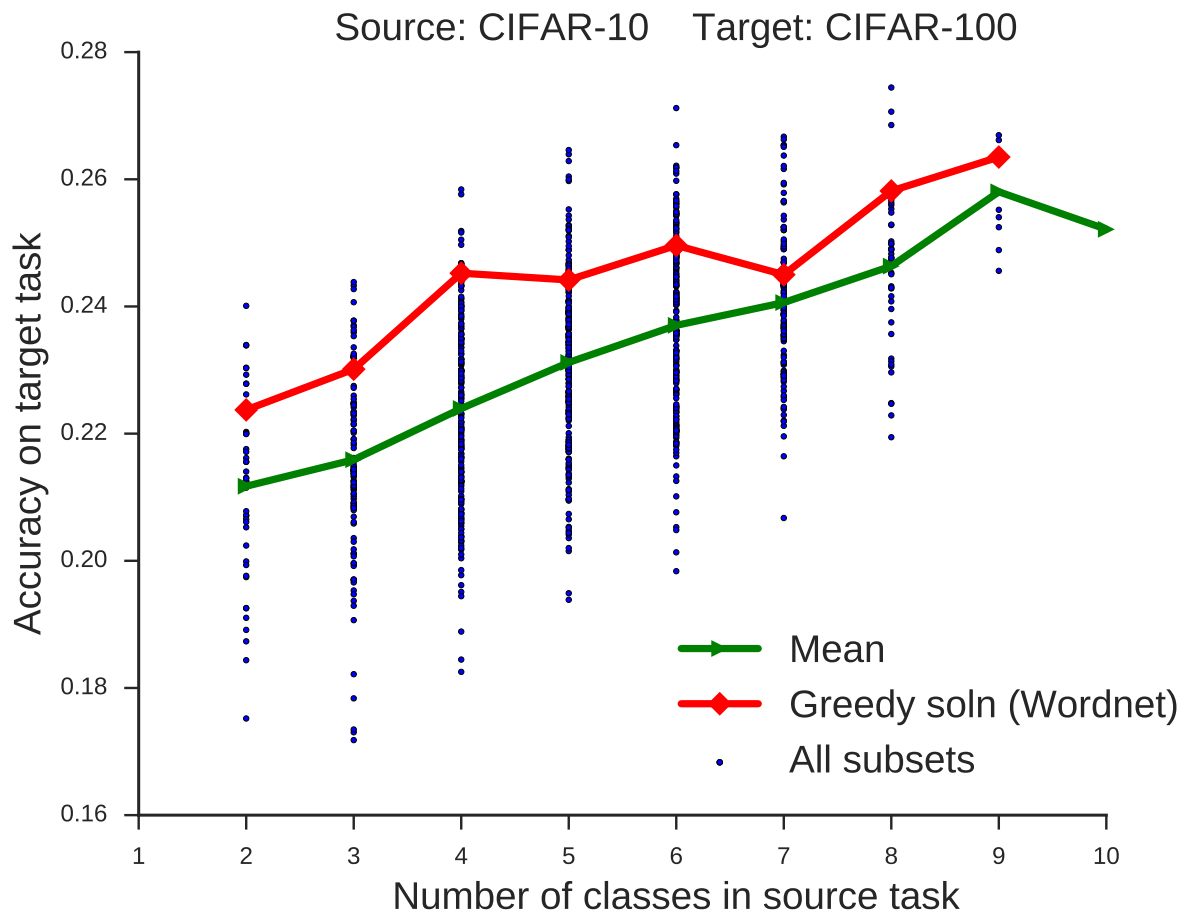

Figure 6. The Optimal SOC obtained using the greedy method (red) outperforms the baseline (green) for most of the cases. The submodular function is computed using similarity matrix.

  many of the experiments, the optimal SOC having 70-80% classes gives better performance on the target task compared to using all of the classes. This behavior can be exploited in generating better pre-trained models for popular datasets such as ImageNet.

- Optimal SOC: In all our experiments we outperformed the random subset baseline in the upper mid-range (40-80%). In subsets with 90% or more elements, there is significant overlap among the different subsets. Thus it is hard for one subset to perform significantly better than the others. The small subsets with 20 or 30 % elements have a diverse set of classes without having pairs of similar ones. Since we don't have a good method to individually evaluate the importance of every class (the $\alpha_i$'s are difficult to obtain for large datasets), it is hard to identify the optimal SOC that performs better than the others.

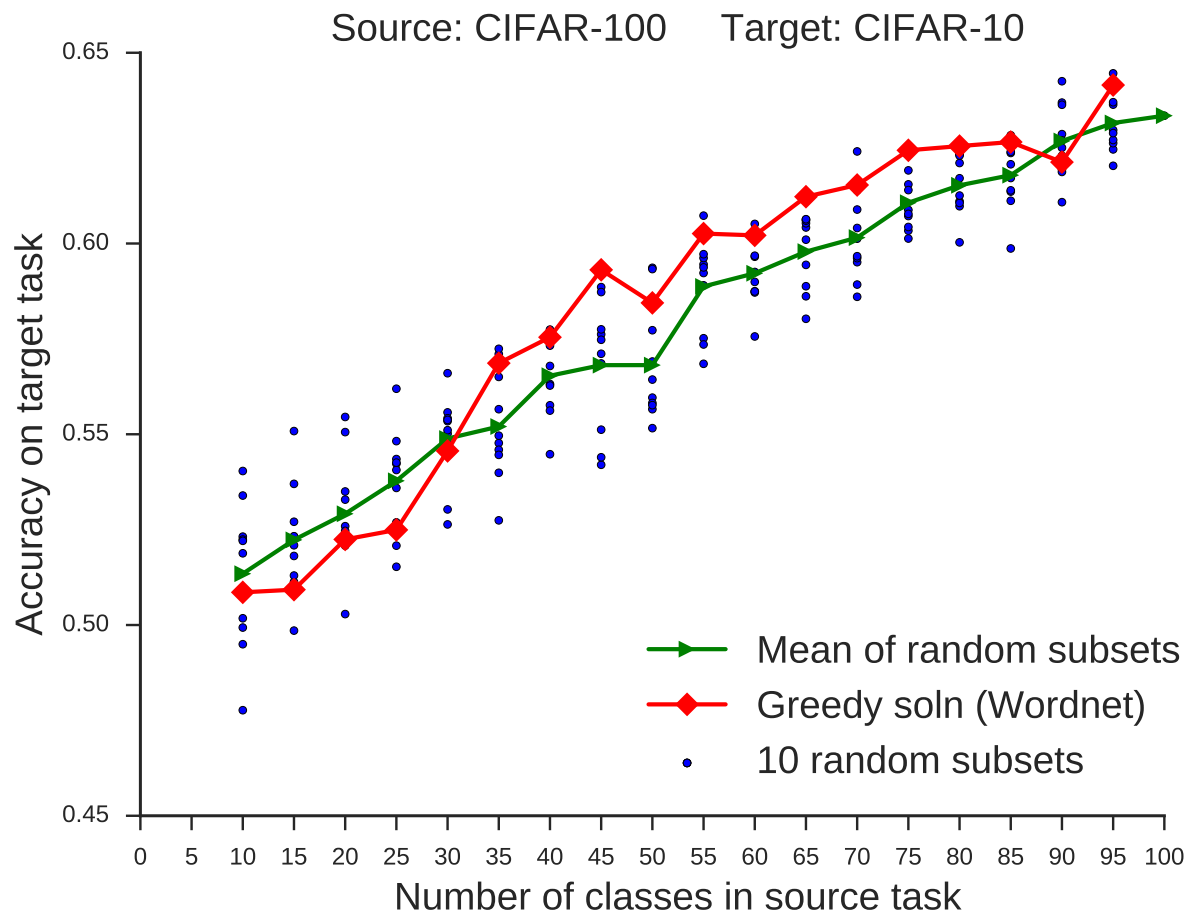

Figure 7. The Optimal SOC obtained using the greedy method (red) outperforms the baseline (green) for most of the cases. The submodular function is computed using similarity matrix. We use 181 random subsets to generate the baseline.

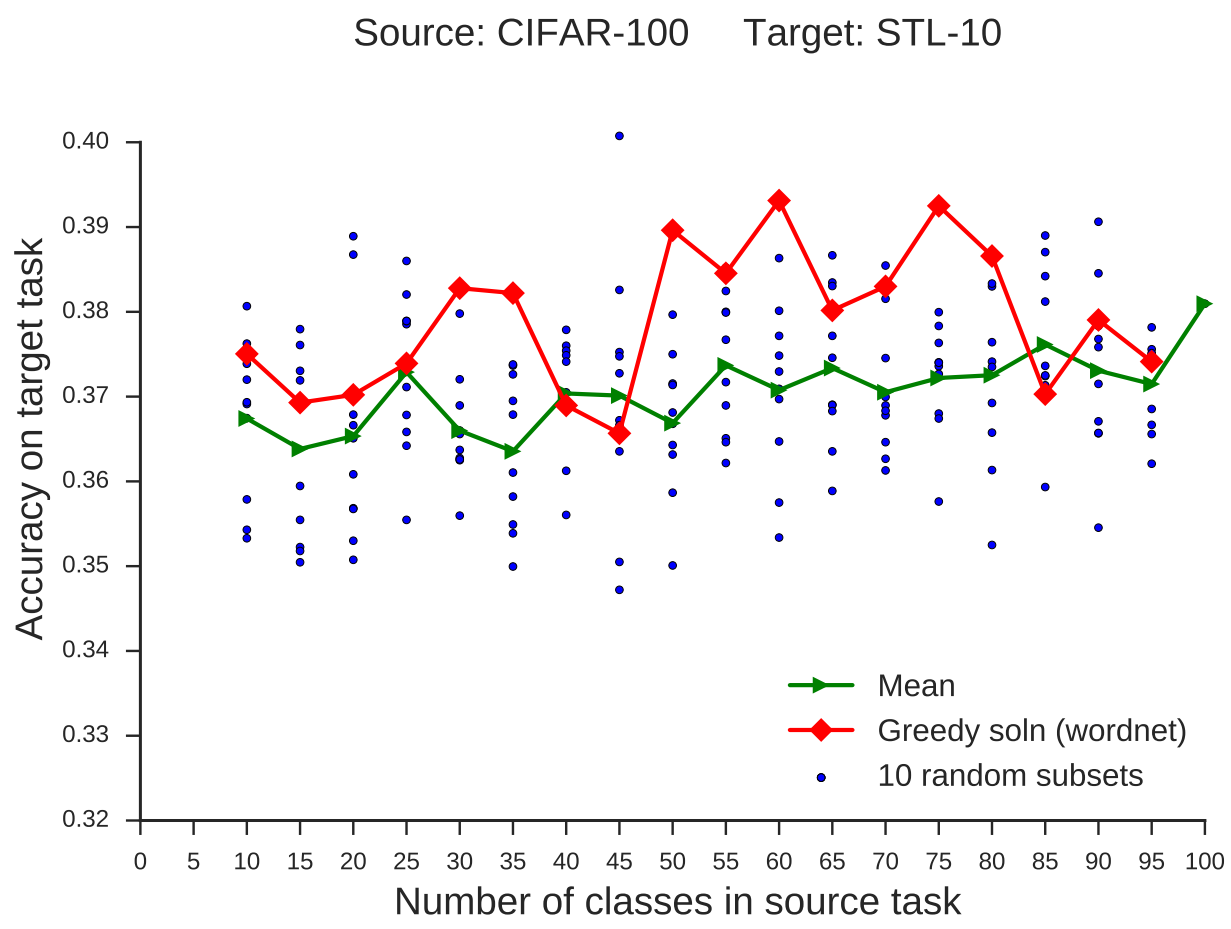

Figure 8. The Optimal SOC obtained using the greedy method (red) outperforms the baseline (green) for most of the cases. The submodular function is computed using similarity matrix. We use 181 random subsets to generate the baseline. Although, the function is mostly monotonically non-decreasing, there is not much gain by adding more classes. This could be due to the domain difference between CIFAR-100 and STL-10 images.

- Object class diversity: In most or all our experiments, we observed that the optimal SOC avoids having two classes that are similar to each other. This is evident from our choice of using $\beta_{ij} = -S(c_i, c_j)$ leading to optimal SOC performing better than the random ones. In all the experiments, the source and target tasks do not share the same classes. Note that CIFAR-10 and CIFAR-100 do not share the same object classes.

In this paper, we wanted to investigate if submodularity can play a role in explaining the performance of generic feature extractors on target tasks. There are many future avenues to explore: (1) the use of more general submodular functions involving higher order functions, (2) investigating the use of non-monotonous submodular functions and non-greedy strategies for finding the optimal SOC and (3) globally optimal approaches for subset selection problems. We studied small and mid-scale datasets and performed a careful analysis of the modeling by probing several SOCs in the source tasks. We are currently looking at computationally efficient ways of using the proposed method for identifying optimal subsets in larger datasets such as ImageNet.

## APPENDIX

## A. Proof for lemma 1

*Proof.* Let us consider the addition of an element $c_i \notin B$ to $B$ where $B \subseteq \mathcal{C}$. We have:

$$F(B \cup c_i) = F(B) + \alpha_i + \sum_{c_j \in B \setminus c_i} \beta_{ij} \tag{9}$$

If $\alpha_i \geq -\sum_{c_j \in \mathcal{C} \setminus c_i} \beta_{ij}$ and $\beta_{ij} \leq 0$, we have

$$\alpha_i \geq -\sum_{c_j \in B, i \neq j} \beta_{ij}, \forall B \subseteq \mathcal{C} \tag{10}$$

From Equation 9, we have:

$$F(B \cup c_i) \geq F(B) \tag{11}$$

By adding newer elements iteratively, we can show the following for all $A, B \subseteq \mathcal{C}$ and $B \subseteq A$, we have $F(B) \leq F(A)$. □

## References

[1] P. Agrawal, J. Carreira, and J. Malik. Learning to see by moving. In *CVPR*, 2015. 3

[2] P. Agrawal, R. Girshick, and J. Malik. Analyzing the performance of multilayer neural networks for object recognition. In *ECCV*, 2014. 3

[3] Z. Akata, F. Perronnin, Z. Harchaoui, and C. Schmid. Label-embedding for image classification. *IEEE Trans. Pattern Anal. Mach. Intell.*, 2016. 5

[4] M. Aubry and B. C. Russell. Understanding deep features with computer-generated imagery. In *ICCV*, 2015. 3

[5] H. Azizpour, A. Razavian, J. Sullivan, A. Maki, and S. Carlsson. From generic to specific deep representations for visual recognition. In *CVPR Workshops*, 2015. 3

[6] A. Bakry, M. Elhoseiny, T. El-Gaaly, and A. Elgammal. Digging deep into the layers of cnns: In search of how cnns achieve view invariance. In *arXiv preprint arXiv:1508.01983*, 2015. 3

[7] S. Bondugula, V. Manjunatha, L. S. Davis, and D. S. Doermann. SHOE: sibling hashing with output embeddings. In *ACM MM*, 2015. 5

[8] E. Boros and P. L. Hammer. Pseudo-boolean optimization. *Discrete Appl. Math.*, 123(1-3):155–225, 2002. 4

[9] A. Coates, A. Y. Ng, and H. Lee. An analysis of single-layer networks in unsupervised feature learning. In *AISTATS*, pages 215–223, 2011. 5

[10] C. Doersch, A. Gupta, and A. A. Efros. Unsupervised visual representation learning by context prediction. In *ICCV*, 2015. 3

[11] A. Dosovitskiy and T. Brox. Inverting convolutional networks with convolutional networks. In *arXiv preprint arXiv:1506.02753*, 2015. 3

[12] R. Girshick, J. Donahue, T. Darrell, and J. Malik. Rich feature hierarchies for accurate object detection and semantic segmentation. In *CVPR*, 2014. 3

[13] P. Grother. Nist special database 19 handprinted forms and characters database. 1995. 1, 5

[14] C. Guestrin, A. Krause, and A. P. Singh. Near-optimal sensor placements in gaussian processes: Theory, efficient algorithms and empirical studies. *Journal of Machine Learning Research*, page 235284, 2008. 3

[15] K. He, X. Zhang, S. Ren, and J. Sun. Deep residual learning for image recognition. In *CVPR*, 2016. 3

[16] M. Huh, P. Agrawal, and A. A. Efros. What makes imagenet good for transfer learning? In *arXiv:1608.08614*, 2016. 3

[17] D. Jayaraman and K. Grauman. Learning image representations tied to ego-motion. In *ICCV*, 2015. 3

[18] D. Kempe, J. Kleinberg, and E. Tardos. Maximizing the spread of influence through a social network. In *KDD*, 2003. 3

[19] A. Krizhevsky. Learning multiple layers of features from tiny images. 2009. 5

[20] A. Krizhevsky, I. Sutskever, and G. E. Hinton. Imagenet classification with deep convolutional neural networks. In *NIPS*, 2012. 1, 3

[21] H. Larochelle, D. Erhan, and Y. Bengio. Zero-data learning of new tasks. In *AAAI*, 2008. 5

[22] Y. LeCun and C. Cortes. MNIST handwritten digit database. 2010. 5

[23] J. Leskovec, A. Krause, C. Guestrin, C. Faloutsos, J. VanBriesen, and N. Glance. Cost-effective outbreak detection in networks. In *The ACM SIGKDD Conference on Knowledge Discovery and Data Mining*, page 420429, 2007. 3

[24] D. Lin. An information-theoretic definition of similarity. In *ICML*, 1998. 5

[25] H. Lin and J. Bilmes. Word alignment via submodular maximization over matroids. In *The 49th Annual Meeting of the Association for Computational Linguistics: Human Language Technologies - Short Papers*, page 170175, 2011. 3

[26] M. Y. Liu, O. Tuzel, S. Ramalingam, and R. Chellappa. Entropy rate clustering: Cluster analysis via maximizing a submodular function subject to a matroid constraint. *IEEE Transaction on Pattern Analysis and Machine Intelligence (TPAMI)*, 2013. 3

[27] S. Mallat. Understanding deep convolutional networks. In *CoRR abs/1601.04920*, 2016. 3

[28] G. L. Nemhauser, L. A. Wolsey, and M. L. Fisher. An analysis of the approximations for maximizing submodular set functions. *Mathematical Programming*, pages 265–294, 1978. 4, 5

[29] M. Noroozi and F. Paolo. Unsupervised learning of visual representations by solving jigsaw puzzles. In *arXiv preprint arXiv:1603.09246v2*, 2016. 3

[30] B. Nushi, A. Singla, A. Krause, and D. Kossmann. Learning and feature selection under budget constraints in crowdsourcing. In *HCOMP*, 2016. 3

[31] D. Pathak, P. Krahenbuhl, J. Donahue, T. Darrell, and A. Efros. Context encoders: Feature learning by inpainting. In *CVPR*, 2016. 3

[32] A. Prasad, S. Jegelka, and D. Batra. Submodular meets structured: Finding diverse subsets in exponentially-large structured item sets. In *Neural Information Processing Systems (NIPS)*, 2014. 3

[33] A. S. Razavian, H. Azizpour, J. Sullivan, and S. Carlsson. CNN features off-the-shelf: An astounding baseline for recognition. In *CVPR*, 2014. 3

[34] O. Russakovsky, J. Deng, H. Su, J. Krause, S. Satheesh, S. Ma, Z. Huang, A. Karpathy, A. Khosla, M. Bernstein, A. C. Berg, and L. Fei-Fei. Imagenet large scale visual recognition challenge. *IJCV*, 2015. 3

[35] K. Simonyan, A. Vedaldi, and A. Zisserman. Deep inside convolutional networks: Visualising image classification models and saliency maps. In *ICLR Workshop*, 2014. 3

[36] K. Simonyan and A. Zisserman. Very deep convolutional networks for large-scale image recognition. In *ICLR*, 2015. 1, 3

[37] C. Szegedy, W. Zaremba, I. Sutskever, J. Bruna, D. Erhan, I. Goodfellow, and R. Fergus. Intriguing properties of neural networks. In *ICLR*, 2014. 3

[38] J. Wang, Z. Zhang, V. Premachandran, and A. Yuille. Discovering internal representations from object-cnns using population encoding. In *arXiv preprint arXiv:1511.06855*, 2015. 3

[39] D. Wei, B. Zhou, A. Torrabla, and W. Freeman. Understanding intra-class knowledge inside cnn. In *arXiv preprint arXiv:1507.02379*, 2015. 3

[40] K. Weinberger and O. Chapelle. Large margin taxonomy embedding with an application to document categorization. In *NIPS*, 2008. 5

[41] X.Wang and A. Gupta. Unsupervised learning of visual representations using videos. In *ICCV*, 2015. 3

[42] J. Yosinski, J. Clune, Y. Bengio, and H. Lipson. How transferable are features in deep neural networks? In *Neural Information Processing Systems*, 2014. 3

[43] M. D. Zeiler and R. Fergus. Visualizing and understanding convolutional networks. In *ECCV*, 2014. 3

[44] R. Zhang, P. Isola, and A. Efros. Colorful image colorization. In *ECCV*, 2016. 3

[45] B. Zhou, A. Khosla, A. Lapedriza, A. Oliva, and A. Torralba. Object detectors emerge in deep scene cnns. In *ICLR*, 2015. 3

# Supplementary Material for Class Subset Selection for Transfer Learning using Submodularity

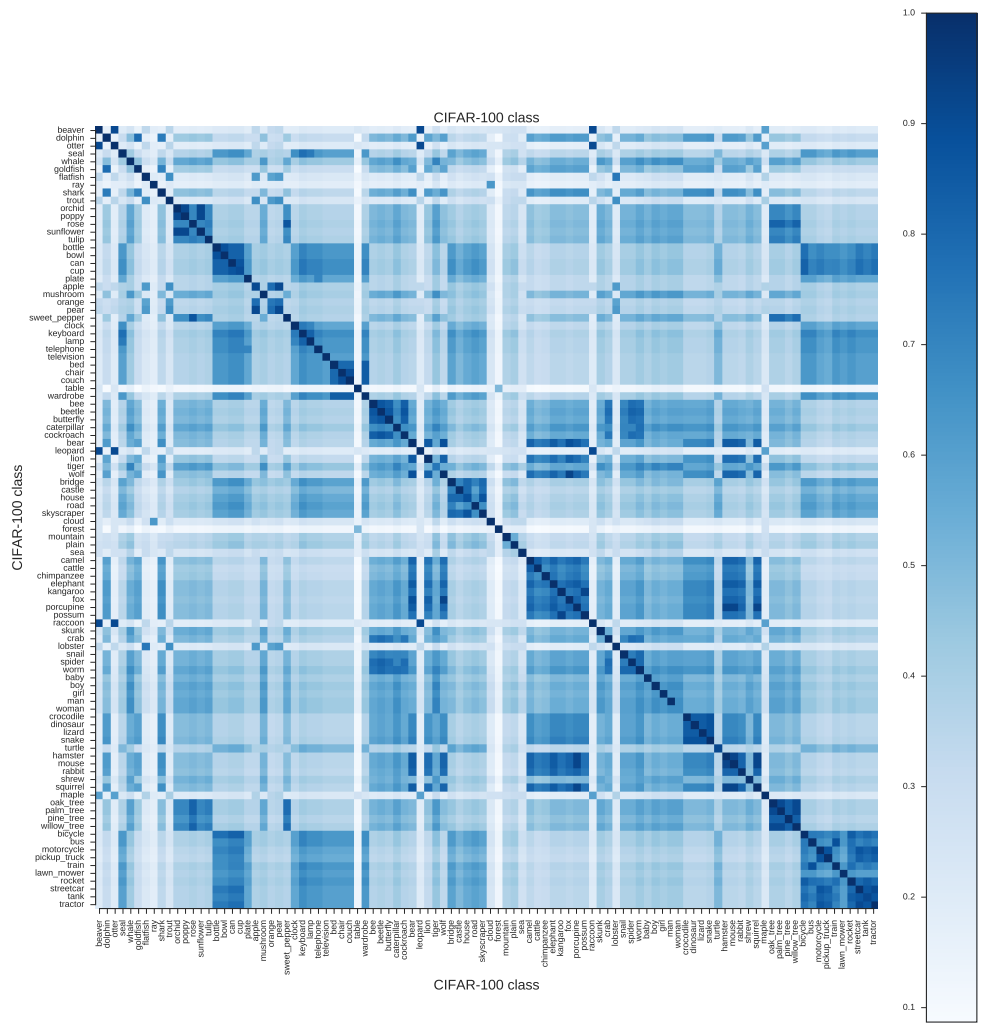

Figure 1. Wordnet similarity matrix for the classes in CIFAR 100. Dark blue entries denote pairs of classes with high similarity. Image best viewed at high resolution.

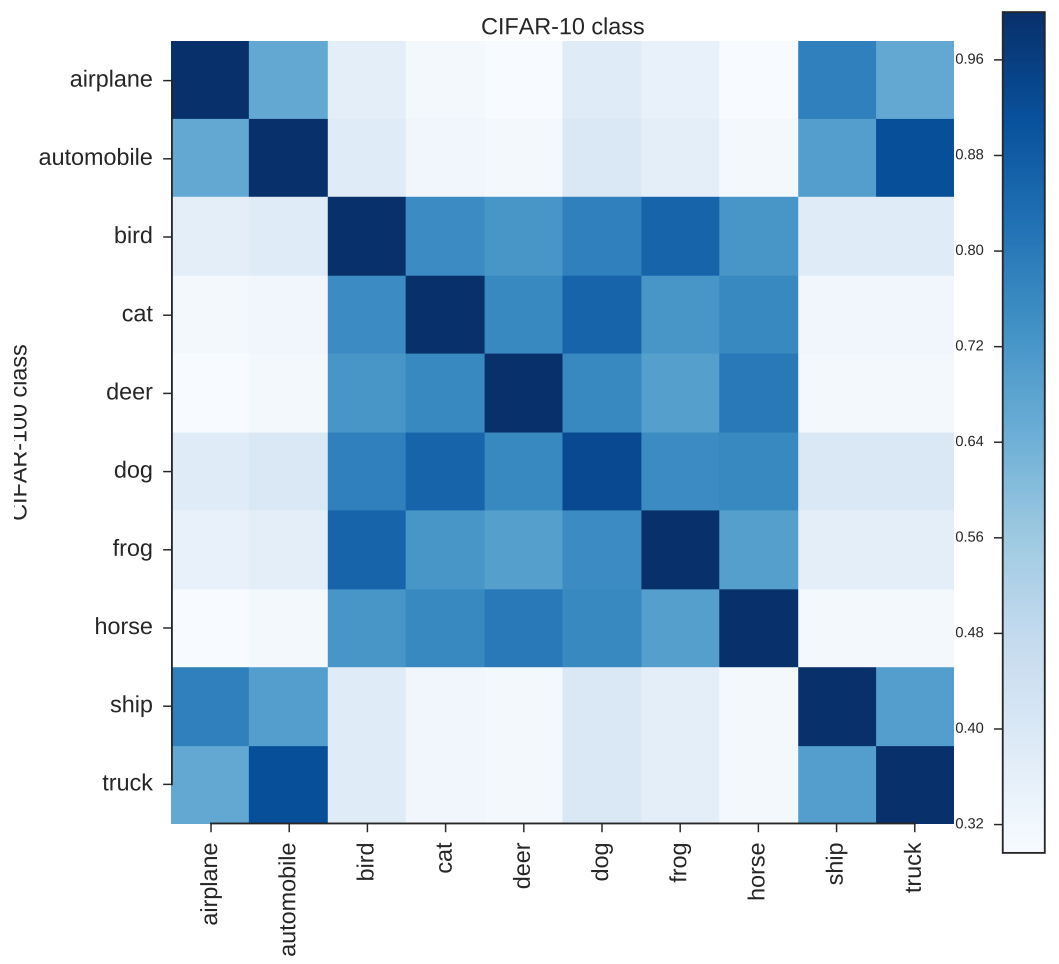

Figure 2. Wordnet similarity matrix for the classes in CIFAR 10. Note that animals are similar to other animals, but are not similar to inanimate objects such as airplane and automobile.

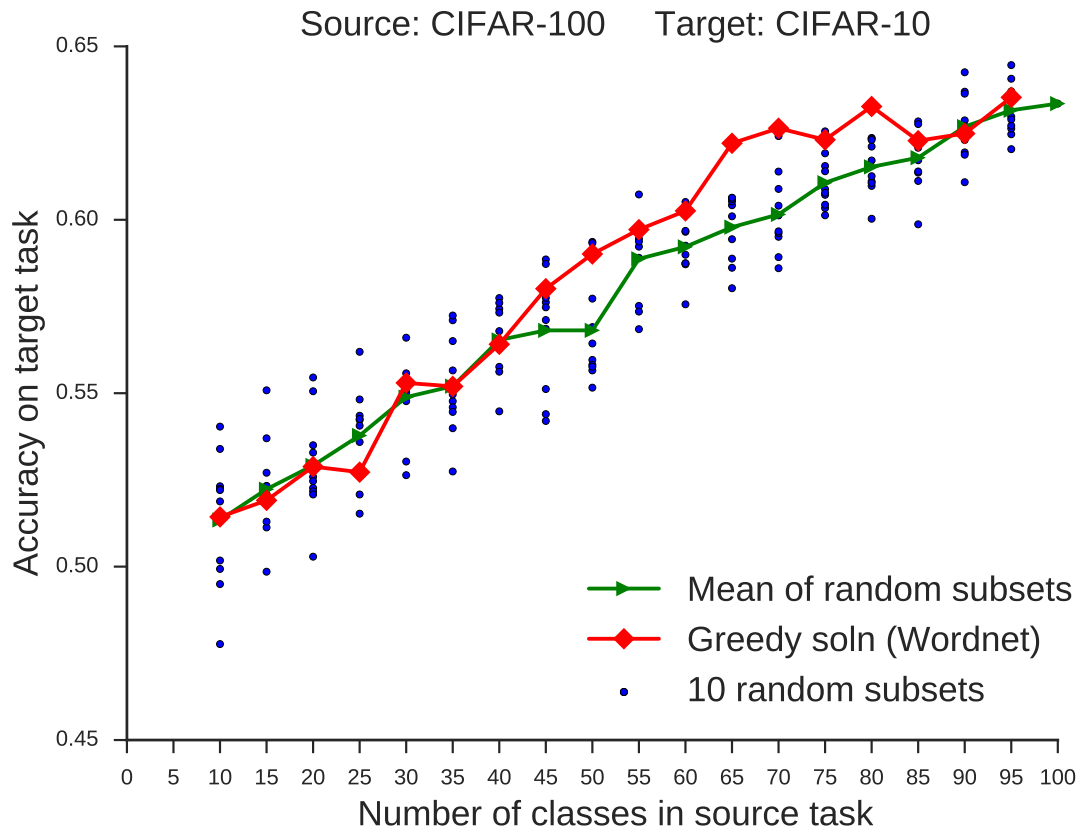

Figure 3. Run 1 of 3. For 3 different runs, the Optimal SOC obtained using the greedy method (red) outperforms the baseline (green) for most of the cases. The submodular function is computed using similarity matrix.

## 1. Wordnet Similarity

In Section 5.1 of the main paper, we used Wordnet (Lin) similarity to compute $\beta_{ij}$ for each pair of classes. In Figures 1 and 2, we visualize this similarity as a confusion matrix for CIFAR-100 and CIFAR-10 respectively. Note that visually similar classes (such as maple, oak_tree, palm_tree, pine_tree) are also similar to each other in the Wordnet space.

## 2. Multiple Runs of CIFAR100-CIFAR10 Experiment

In Figures 3–5, we demonstrate that our experiments are not sensitive to the randomness inherent in training CNNs by plotting the results from multiple runs of the CIFAR100-CIFAR10 experiment with different random seeds. Observe that our OPT-SOC solution consistently outperforms the mean baseline in the 40%–80% range, as we observed in the Discussion section of the main paper.

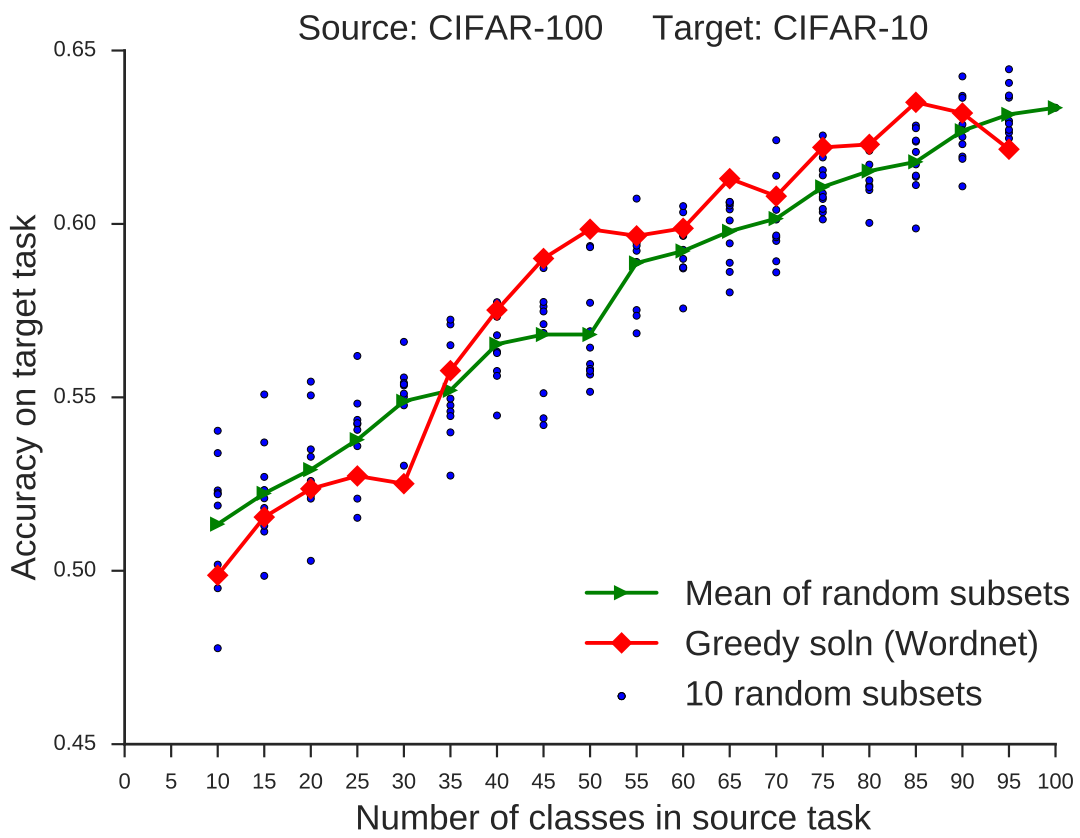

Figure 4. Run 2 of 3. For 3 different runs, the Optimal SOC obtained using the greedy method (red) outperforms the baseline (green) for most of the cases. The submodular function is computed using similarity matrix.

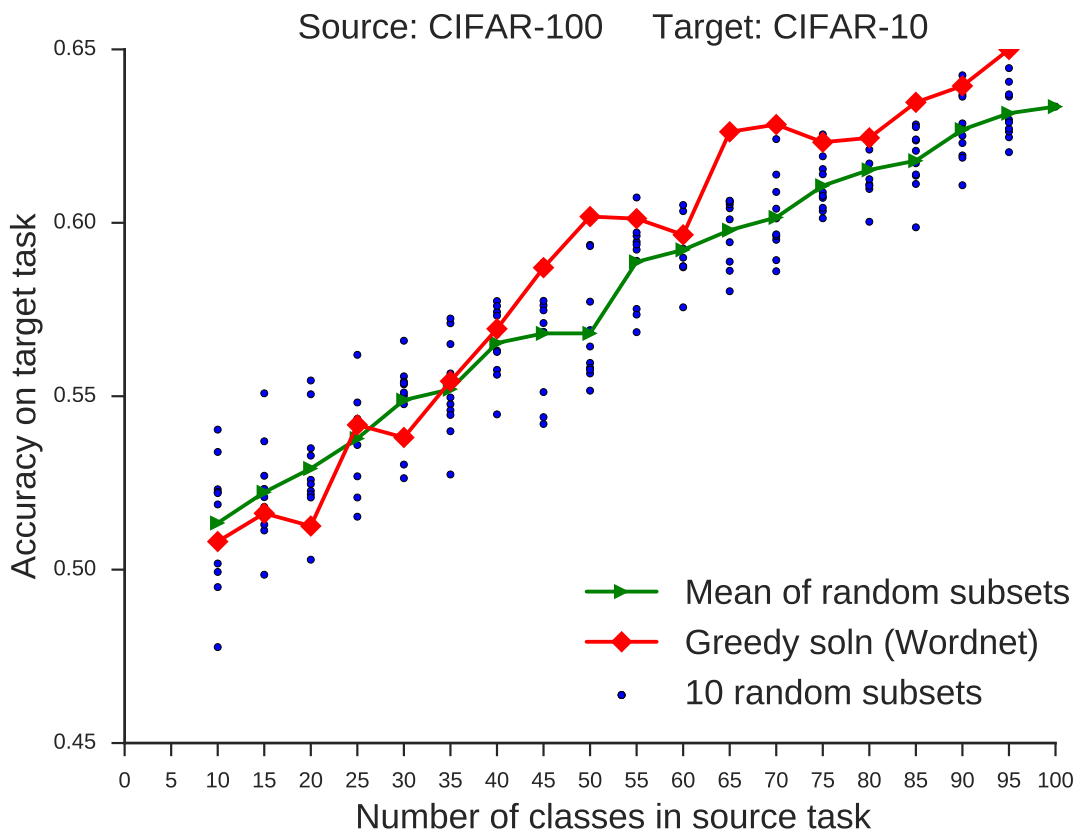

Figure 5. Run 3 of 3. For 3 different runs, the Optimal SOC obtained using the greedy method (red) outperforms the baseline (green) for most of the cases. The submodular function is computed using similarity matrix.